\documentclass[conference]{IEEEtran}
\IEEEoverridecommandlockouts
\usepackage{cite}
\usepackage{amsmath,amssymb,amsfonts}
\usepackage{algorithmic}
\usepackage{url}
\usepackage{graphicx}
\usepackage{textcomp}
\usepackage{xcolor}
\usepackage{colortbl}
\usepackage{hyperref}
\hypersetup{
  colorlinks=true,
  urlcolor=blue,      
  linkcolor=black,    
  citecolor=black     
}
\usepackage{fontawesome5}
\def\BibTeX{{\rm B\kern-.05em{\sc i\kern-.025em b}\kern-.08em
    T\kern-.1667em\lower.7ex\hbox{E}\kern-.125emX}}

\usepackage{amsfonts} 
\usepackage{amsmath}  
\usepackage[mathscr]{euscript} 
\usepackage{tikz}
\newcommand{\circled}[1]{\tikz[baseline=(char.base)]{\node[shape=circle,draw,inner sep=1pt] (char) {#1};}}

\newcommand\Nor{\mathcal{N}}
\newcommand{\ttheta}{\mathbf{\theta}}
\newcommand\Ent{\mathcal{H}}

\newcommand\Esp{\mathbb{E}}
\DeclareMathOperator*{\argmax}{arg\,max}

\newcommand{\data}{\mathcal{D}}

\begin{document}

\title{Active Exploration in Bayesian Model-based \\Reinforcement Learning for Robot Manipulation\\
\thanks{This work was supported by a DGA scholarship, DGA project T45\_23R, by grants PID2024-159284NB-I00 and PID2024-158322OB-I00 funded by MCIN/AEI/10.13039/501100011033 and ERDF.}
}

\author{
\IEEEauthorblockN{Carlos Plou}
\IEEEauthorblockA{
\textit{DIIS-I3A} \\
\textit{Univ. de Zaragoza}\\
Zaragoza, Spain \\
c.plou@unizar.es}
\and
\IEEEauthorblockN{Ana C. Murillo}
\IEEEauthorblockA{
\textit{DIIS-I3A} \\
\textit{Univ. de Zaragoza}\\
Zaragoza, Spain \\
acm@unizar.es}
\and
\IEEEauthorblockN{Ruben Martinez-Cantin}
\IEEEauthorblockA{
\textit{DIIS-I3A} \\
\textit{Univ. de Zaragoza}\\
Zaragoza, Spain \\
rmcantin@unizar.es}

}

\maketitle

\begin{abstract}
Efficiently tackling multiple tasks within complex environments, such as those found in robot manipulation, remains an ongoing challenge in robotics and an opportunity for data-driven solutions such as reinforcement learning (RL). The high cost of data gathering in robotics encourages data-efficient approaches such as model-based RL, which builds a dynamic model of the robot to conduct most policy learning on cheaper simulations, while enabling data reuse and transfer between tasks.
In this work, we employ Bayesian deep learning (BDL) with a double purpose. First, we increase the model quality by improving its calibration. Second, we improve data efficiency by performing active learning of the dynamic model during a preliminary exploration phase, actively estimating the information gain of each transition and using it as the exploration reward. We compare several BDL methods in a realistic robot manipulation setup.
Our experiments show that our approach matches the quality of relevant alternatives with much lower robot execution requirements and, unlike previous studies validated only on standard RL environments, we tackle robotic arm end-tasks on a complete manipulator simulator such as \texttt{CoppeliaSim}. Code is available at \href{https://github.com/ropertunizar/BayesianActiveExploration}{\faGithub\ \texttt{BayesianActiveExploration}}.
\end{abstract}


\section{INTRODUCTION}

Flexible robotic manipulation is a  crucial application domain for robotics, offering potential benefits to various industries, including manufacturing, supply chain, and healthcare \cite{han2023survey}.
In this paper, we tackle robotic manipulation problems through the lens of reinforcement learning (RL), which has shown a great potential to acquire complex behaviors from low-level data \cite{ferrandis2024visuotactile}.
In RL, the robot learns a policy that maximizes a reward signal for solving a specific task. Despite the potential of RL, the problem of flexible manipulation introduces various challenges, including the high complexity of manipulation tasks and the need to efficiently tackle several tasks and transfer or reuse knowledge.

\begin{figure}[!t]
    \centering
\includegraphics[width=\linewidth]{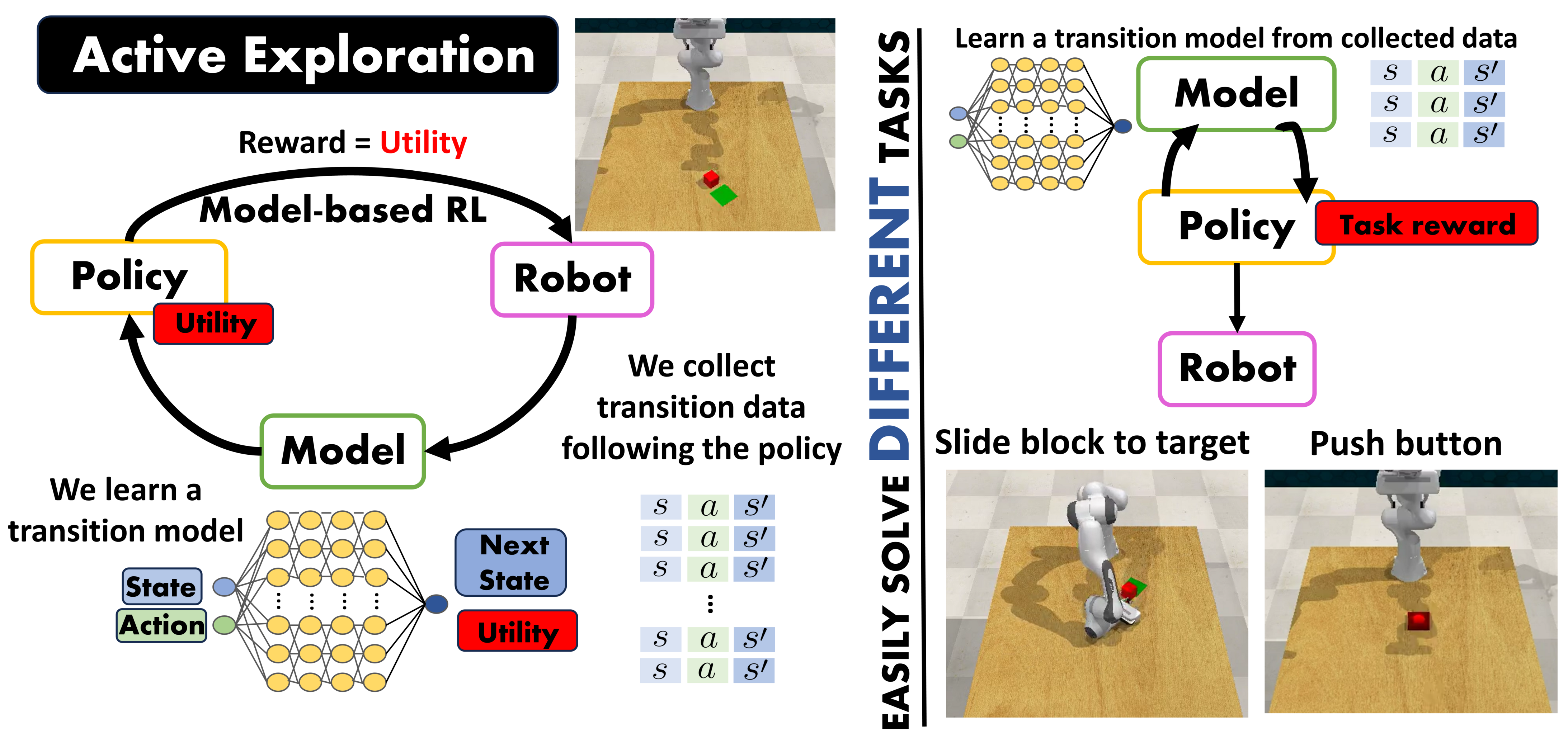}
    \caption{Overview of the active exploration problem through Bayesian model-based RL. The Bayesian model is responsible for predicting both the next state distribution and its degree of novelty. Lastly, we exploit the knowledge acquired during exploration to solve different tasks.}
    \label{fig:Teaser}
\end{figure}

RL strategies can be grouped in: (1) model-based approaches in which the policy is built from simulated episodes, using a model that has been previously learnt to emulate the dynamics of the robot and the environment, and (2) model-free strategies in which the policy is directly built from episode execution in the robotic platform. Model-based RL (MBRL) is more sample-efficient than model-free methods, requiring less real robot interaction to learn policies, thereby reducing costs~\cite{chua2018deep}. Additionally, the dynamic model is task-agnostic and can be reused across tasks in the same environment.

Developing a model to simulate dynamics requires comprehensive and informative data to ensure effective model's training and, thus, policy creation. Active learning appears as a suitable tool because it selectively chooses data points that offer the most information or learning progress, achieving comparable or better performance with reduced training. Similarly, in this work, we guide exploration toward the most unknown or informative regions of the state space to learn more about them. The main challenge is defining a criterion (utility) that measures the information gain from visiting a region, requiring a way to capture the model's uncertainty and the value of new observations. One approach for incorporating uncertainty is through Bayesian inference, such as Bayesian deep learning (BDL). In this paper, we make our dynamic model Bayesian using BDL and, hence, we are able to compute the information gain utility of the exploration problem.

Our main contributions are: 1)
An experimental \textbf{demonstration} of the potential of \textbf{active learning, exploration and Bayesian deep learning (BDL)} in the context of model-based reinforcement learning (MBRL) for \textbf{robot manipulation}
using \texttt{CoppeliaSim}. 2) We are the first to validate \textbf{Monte Carlo dropout} and \textbf{Laplace approximation} for model building in MBRL, showing comparable performance to standard \emph{deep ensembles} with much lower training time and memory footprint.
3) A \textbf{novel information-based utility function} for active exploration based on relative Shannon entropy, which is more accurate and better calibrated with respect to model errors.
Figure~\ref{fig:Teaser} shows an overview of our work.

\section{RELATED WORK}

\textbf{Model-based Reinforcement Learning:}
Among RL directions, MBRL stands out for its sample efficiency. The dynamics model can be non-parametric, such as Gaussian processes \cite{deisenroth2011pilco}, or parametric, such as neural networks \cite{chua2018deep}. An alternative involves world models, which leverage generative models to hallucinate the environment and build latent spaces where the agent can plan or learn the control policy \cite{hafner2025mastering, hansen2024tdmpc2}. However, working on those latent spaces is complex, subject to unexpected hallucinations and where the objective might not be completely aligned with the task objective \cite{ghugaresimplifying}. Instead, in this work we focus on a more grounded state-space representation based on the actual robot and environment dynamics \cite{sancaktar2022curious}, enabling fine-grained actions and a hierarchy of models for lifelong learning \cite{fu2022model}.

\textbf{Bayesian deep learning:}
To our knowledge, most MBRL with Bayesian neural networks rely on deep ensembles \cite{lakshminarayanan2017simple} which states to train N models with different architectures, hyperparameters, or initial weights \cite{chua2018deep,sancaktar2022curious,MAX}, while few works use mean-field variational inference within RL or MBRL \cite{houthooft2016vime,fu2022model}. While deep ensembles are shown to have good performance \cite{gustafsson2020evaluating, wang2024credal}, they are computationally expensive method due to the need to train multiple networks. In contrast, variational methods are known to have many accuracy and calibration issues \cite{pmlr-v108-farquhar20a}. On the other hand, Monte Carlo (MC) dropout is specific type of variational inference that approximates the posterior distribution through sampling M forward passes, while maintaining dropout layers active during inference \cite{gal2016dropout}, again requiring training the network only once while having reasonable performance \cite{kendall2017uncertainties}. Differently, the Laplace approximation tries to find a Gaussian approximation to the posterior distribution of the weights \cite{mackay1992}, which has great advantages for robotics applications and provides better calibrations. Although seminal, the Laplace approximation has recently been made much more tractable and accurate \cite{daxberger2021laplace, cinquin2024fsplaplace, roy2024reparameterization}.

\textbf{Exploration and Active Learning in DRL:}
Most of the exploration algorithms in RL are reactive exploration methods. These methods rely on chance discoveries to drive exploration \cite{houthooft2016vime}.
In contrast, a new trend is active exploration which exploits the active learning theory \cite{MAX,pathak2017curiosity,sukhija2023opax}.
In supervised learning, active learning iteratively selects the most informative instances from an unlabeled dataset, which can be easily adapted to RL. Previous work has addressed exploration for Transition Query Reinforcement Learning where all transitions are available for querying directly \cite{mehta2021experimental}. However, in the general case in RL, most transitions are only available once the agent has reached a certain state, forcing to develop full active exploration policies instead of querying methods.
This approach requires a novelty or information-based measure, whose existing formulations include visitation count, prediction error \cite{pathak2017curiosity}, learning progress and diversity in the visited states. Other approaches leverage deep ensembles technique to compute the utility as either the variance \cite{pathak2019self} or the Jensen-Rényi Divergence \cite{MAX} among samples. In contrast, we measure utility leveraging the entropy of the predictive distribution computed by Laplace Approximation or MC-dropout. While we explore for active model learning in a task-agnostic fashion, other methods explore to solve tasks using rewards that combine model accuracy and task utility \cite{ghugaresimplifying} or using optimism in the face of uncertainty \cite{fu2022model, sukhija2025maxinforl}, sometimes adding safety constraints \cite{as2025actsafe}. However, by solving the specific reward during training, the model is biased and might not be able to generalize to other tasks, requiring complex hierarchical models and curriculum learning \cite{fu2022model}.

\section{BAYESIAN MODEL-BASED RL}
\label{sec:mbrl}
\subsection{Background}
Let the problem be represented as a Markov decision process (MDP) tuple $(\mathcal{S}, \mathcal{A}, t^*, R, \rho_0)$ where \circled{1} $\mathcal{S}$ is the state space, \circled{2} $\mathcal{A}$ is the action space, \circled{3} $t^* = p_*(s'|s, a)$ 
is the unknown transition function, i.e. the probability of reaching next state $s'$ given the current state $s$ and taking action $a$, \circled{4} $R: \mathcal{S} \times \mathcal{A} \rightarrow \mathbb{R}$ is the reward function and \circled{5} $\rho_0$ is the initial state distribution. Without loss of generality, we assume continuous state and action spaces $(\mathcal{S}, \mathcal{A})$.

In MBRL we build a probabilistic model to predict the distribution of the actual transition function, i.e., $p_{*}(s'| s, a)$.
 We will regard this model as a neural network with weights $\ttheta$. This neural network can output the parameters of a Gaussian distribution $(\mu_\ttheta, \sigma_\ttheta)$ so as to predict the probability distribution $p(s'| s, a, \ttheta)$. In mathematical terms, 
 \begin{equation}
    p (s' | s, a, \ttheta) \sim \Nor (s' | \mu_\ttheta(s, a), \sigma_\ttheta(s, a)):=\Nor_\ttheta.
 \end{equation}
This model can be trained from a replay buffer of collected transitions $\mathcal{D}$ -i.e., sequences of states and actions as inputs $\{(s_{i}, a_{i})\}_{i=1}^{n}$ and next states as outcomes or labels $\{s'_{i}\}_{i=1}^{n}$- to 
find the \textit{maximum a posteriori (MAP)} of their weights,
\begin{equation}\label{MAP}
    \ttheta_{MAP} = \argmax_{\ttheta} \log p(\mathcal{D}|\ttheta)p(\ttheta),
\end{equation}
where, first, $p(\mathcal{D}|\ttheta)$ is the \textit{likelihood} 
and, second, the \textit{prior} distribution of the weights is $p(\ttheta) \sim \Nor(0, 1/\gamma^2)$,
where $\gamma^2$ is the prior precision. As result, we get $\Nor_{\ttheta_{MAP}}$.

However, this strategy lacks proper uncertainty calibration because it captures only the uncertainty associated with the data, called \emph{aleatoric uncertainty}, but it misses any model uncertainty, called \emph{epistemic uncertainty} \cite{kendall2017uncertainties}. A more reliable prediction is to compute a posterior distribution over model parameters $p(\ttheta|\data)$ and compute the transition function by marginalization of those parameters, that is:
\begin{equation}\label{eq:pred}
    p(s'| s, a) = \int_\Theta p(s'|s,a, \ttheta) p(\ttheta|\data) d \ttheta.
\end{equation}

Capturing the epistemic uncertainty lets us determine how much a model knows and, more importantly for exploration, how much it can learn from new data. This equation is intractable, so we rely on the approximate solutions described next.

\subsection{Approximate Bayesian inference}\label{sec:approxbayes}

\textbf{Deep ensembles-.}
This method uses Bayesian model average, where we build an ensemble of $N$ trained models, typically from random initializations and optimized following equation \eqref{MAP}, resulting in a sample approximation of the posterior distribution $\left\{ \ttheta_{1}, \ldots, \ttheta_{N}\right\}$ \cite{lakshminarayanan2017simple}. While the samples are not strictly from the full posterior distribution, starting from random values can capture its multimodality \cite{gustafsson2020evaluating} and the sample disagreement can be considered an approximation of posterior uncertainty \cite{pathak2019self}.
Deep ensembles are highly expensive both in terms on computing power, as the training process needs to be replicated multiple times, and in memory footprint, as it needs to manage and maintain several models at once. Furthermore, as the models need to be trained from random initializations, they cannot be pretrained or fine-tuned, limiting the transfer learning capabilities. 

\textbf{Variation inference-.} Mean-field variational inference as used in RL \cite{fu2022model,houthooft2016vime}, assumes an isotropic Gaussian distribution for $q(\theta)\sim \mathcal{N}(\theta|\mu,\mathbf{I}\sigma)$. The parameters of the approximate distribution $q(\theta)$ are optimized by minimizing the KL-divergence between the approximate distribution and the true posterior $D(q||p)$. Mean-field variational inference with Gaussian approximation suffers from the soap-bubble effect, reducing the predictive performance as most samples fall in a ring \cite{pmlr-v108-farquhar20a}. Furthermore, VI methods are very sensitive to calibration and configuration.

\textbf{Monte Carlo dropout-.} While theoretically being a special case of VI, in practice it is quite different, resulting in a much better performance. MC droput approximates the posterior as the sample distribution of $N$ forward passes during
 inference time, maintaining active the random dropout layers \cite{gal2016dropout}.

\textbf{Laplace approximation-.} Trying to cover the predictive distribution \eqref{eq:pred}  in a more reliable way, we may approximate the posterior distribution 
of the weights through Laplace approximation \cite{mackay1992} as a Gaussian distribution,
\begin{equation}\label{eq:LapApp}
   p(\ttheta|\mathcal{D}) \approx \Nor(\ttheta |\ttheta_{MAP}, \mathbf{H}^{-1}),
\end{equation}
where $\mathbf{H}=\nabla^{2}_{\ttheta} \log p(\data|\ttheta)|_{\ttheta_{MAP}}$
and, subsequently, translate this uncertainty into the predictive variable. A first-order Taylor approximation (linearization) allows us to get a Gaussian predictive distribution 
\begin{equation}\label{eq:PredLap}
    p(s'| s, a) \approx \Nor(s' | \mu_{\ttheta_{MAP}}, \mathbf{J}^{T}\mathbf{H}^{-1}\mathbf{J} + \text{diag}(\sigma_{\ttheta_{MAP}})),
\end{equation} 
where $[\mu_{\ttheta_{MAP}},\sigma_{\ttheta_{MAP}}] = f_{\ttheta_{MAP}}(s,a)$ is the network prediction and $\mathbf{J}=\nabla_{\ttheta}\mu_{\ttheta}(s, a)|_{\ttheta_{MAP}}$. 
To implement Laplace approximation, we use the recent \emph{Laplace redux} library \cite{daxberger2021laplace}, which offers various approximations tailored for modern deep neural networks. For example, when dealing with larger models, we are obliged to apply subnetwork inference. This technique considers as Bayesian weights only the $n_{sub}$ weights with greatest absolute values as they generate the largest activations
\cite{daxberger2021laplace}. Besides, to translate this uncertainty into the predictive distribution, we use Monte Carlo, that is, we take $N$ samples of the subnetwork weights from \eqref{eq:LapApp} and perform a forward pass per sample,  setting the remaining weights with their MAP estimate value.

In our preliminary experiments, variational inference methods performed poorly compared to other alternatives and proper deep ensembles are too expensive for robotic applications. Instead, we focus on \emph{Laplace approximation} and \emph{MC dropout} which share the following characteristics: computationally and memory efficient, accurate and calibrated predictions and the possibility of using pretrained networks. For the rest of the paper, we focus on state-space Markov models of the actual robot and environment dynamics, though these methods could also apply to world models and latent-space planning.

\section{EXPLORATION FOR MBRL}
In this context, we define the exploration problem as gathering valuable information in the replay buffer $\mathcal{D}$ to learn the most accurate dynamic model $p(s'|s,a)$.  Therefore, we can pose the exploration problem as an active learning problem. However, contrary to query strategies found in classical active learning \cite{mehta2021experimental}, in robotics data cannot be queried independently, since a state is only reachable through a preceding sequence of actions. Our exploration problem can be defined as a MDP $(\mathcal{S}, \mathcal{A}, t^*, u, \rho_0)$, similar to the one presented in Section \ref{sec:mbrl},  replacing the reward task  by a utility function $u$ that measures  the transition novelty. This exploration and posterior model learning are task agnostic and independent of the policy learning process (SAC) that comes afterwards. Hence, both could potentially generalize to any task.

\subsection{Exploration as experimental design}
In our framework, we first set an exploration MDP where the reward term is an experimental design based utility. This utility is based on the expected information gain (IG) obtained by reaching and observing future state-action pairs $(s,a, s')$. 
The \emph{information gain} may be defined as the amount of information gained about a random variable or signal from observing another random variable. In our scenario, the information gain by observing a transition $z = \{s, a, s'\}$ can be defined as the KL-divergence of the model posterior distribution after and before adding the transition data, that is:
\begin{equation}\label{eq:IG}
    IG(s, a, s') = D_{KL}\left(p(\ttheta|\data \cup z) \;\Vert\; p(\ttheta|\data)\right),
\end{equation}
From this expression, the utility of an action at a state is:
\begin{equation}\label{eq:Utility}
    u(s, a) = \int_\Theta \int_{\mathcal{S}}IG(s, a, s')p(s'|s, a, \ttheta) p(\ttheta|\mathcal{D}) d\ttheta ds',
\end{equation}
which can be used to compute the utility of a policy 
\begin{equation}
    U(\pi) = \Esp_{p(\ttheta|\data)}\left[\Esp_{p(s,a|\pi,\ttheta)}[u(s, a)]\right]
\end{equation}
where $p(s,a|\pi,\ttheta)$ is the probability of being in state $s$ and selecting action $a$ by following the policy $\pi$ and the transition model defined by $\ttheta$. Then, we can also define the action utility as the Jensen-Shannon (JS) divergence following the derivation of Shyam et al. \cite{MAX}:
\begin{equation}
    u(s, a) = \Ent_{s'} \left(\Esp_{\ttheta}\left[p(s'|s, a, \ttheta)\right]\right) - \Esp_{\ttheta} \left[\Ent_{s'} \left( p(s'|s, a, \ttheta) \right)\right] 
\end{equation}
note that the first term corresponds to the entropy of $\Esp_{\ttheta}\left[p(s'|s, a, \ttheta)\right]$, which is exactly the predictive posterior as defined in equation \eqref{eq:pred}. Conceptually, it captures the total uncertainty (epistemic and aleatoric) of our model. The second term represents the expected entropy of the predictive distribution, that is, the aleatoric uncertainty. Therefore, the JS divergence is maximized where the epistemic uncertainty is maximized. Intuitively, exploration is driven towards states and actions where uncertainty is high because the model lacks information, not because the data are noisy.

If we have a sample representation of our model $\{\ttheta_i\}_{i=1}^N$, like in the case of deep ensembles of MC-dropout, we can compute the utility as the utility of a mixture of Gaussians:
\begin{equation}
    \label{eq:js_sample}
    u(s, a) = \Ent \left(\frac{1}{N}\sum_{i=1}^{N}\mathcal{N}_{\ttheta_{i}} \right) -\frac{1}{N}\sum_{i=1}^{N} \Ent \left( \mathcal{N}_{\ttheta_{i}} \right). 
\end{equation}
However, the first term is the Shannon entropy of a mixture of Gaussians which does not have analytical solution. Instead, we can use the generalized R\'enyi entropy $\mathcal{H}_\alpha$ with $\alpha=2$ which has a closed form \cite{MAX}. Note that the Shannon entropy is $\mathcal{H} = \lim_{\alpha \rightarrow 1} \mathcal{H}_\alpha$. R\'enyi entropy with $\alpha > 1$ is biased towards high-probability events and ignores low-probability ones.

\subsection{Entropy metric for Gaussian approximation} \label{sec:entropy}
One advantage of the Laplace approximation is that the posterior model is approximated as a Gaussian distribution with a predictive covariance $\Sigma_{LA} = \mathbf{J}^{T}\mathbf{H}^{-1}\mathbf{J} + \sigma_{\ttheta_{MAP}}(s, a))$.  In this way, Laplace approximation enables us to directly tackle (9) and, thus, defining a novel utility measure. First, it is important to note that,
\begin{align*}
    \label{eq:js_gaussian}
    \Ent_{s'} \left(\Esp_{\ttheta}\left[p(s'|s, a, \ttheta)\right]\right) &\stackrel{c}{=} -\log|\Sigma_{LA}|\\
    \Esp_{\ttheta} \left[\Ent_{s'} \left( p(s'|s, a, \ttheta) \right)\right] &\stackrel{c}{=} -\Esp_{\ttheta} \left[\log|\sigma_{\ttheta}(s, a))|\right]
\end{align*}
where $\stackrel{c}{=}$ represents \emph{equal up to a constant term}. In general, the second term does not have analytical form. However, we can assume homoscedastic noise, that is, $\sigma_\ttheta$ is a constant value, independent of $(s, a)$. In practice, our BDL model learns heteroscedastic noise , dependent of the current state and action, but it has negligible variability compared to other terms. Therefore, if we assume homoscedastic noise within our utility function, we can simplify it:
\begin{equation}
    u(s,a) \stackrel{c}{=} \Ent_{s'} \left(\Esp_{\ttheta}\left[p(s'|s, a, \ttheta)\right]\right) \stackrel{c}{=} -\log|\mathbf{J}^{T}\mathbf{H}^{-1}\mathbf{J}|.
\end{equation}
This novel metric, \emph{\textbf{entropy metric}}, can be seen as a particular case of the aforementioned idea that epistemic uncertainty has to be the driving factor in exploration. This metric is also related to other metrics based on disagreement \cite{pathak2019self}. 

\subsection{Entropy metric for sample approximation}
We can also define a metric based purely on the entropy of the predictive model using the sample approximation of MC-dropout or deep ensembles. Starting from \eqref{eq:js_sample}, instead of approximating by Rényi entropy, we can  approximate the mixture of Gaussians $\left\{\mathcal{N}_{\ttheta_{1}}, \ldots, \mathcal{N}_{\ttheta_{N}}  \right\}$ by a new Gaussian $\mathcal{N}_{M}$ with  mean $\mathbf{\mu}_{M} = \frac{1}{N}\sum_{i=1}^{N} \mu_{\ttheta_i}$ and covariance $\Sigma_{M}$ defined as,
\begin{equation}
\Sigma_{M} = \frac{1}{N} \sum_{i=1}^{N} \left( \text{diag}(\sigma_{\ttheta_i}) + \mu_{\ttheta_i}\mu_{\ttheta_i}^T \right) -  \mu_{M}\mu_{M}^T.
\end{equation}
Then, if we also assume homoscedastic noise, we get the entropy metric based on samples:
\begin{equation}
    u(s,a) \stackrel{c}{=} -\log \left| \frac{1}{N} \sum_{i=1}^{N}  \mu_{\ttheta_i}\mu_{\ttheta_i}^T -  \mu_{M}\mu_{M}^T \right|.
\end{equation}

\begin{figure}[!t]
    \centering
\includegraphics[width=
\linewidth]{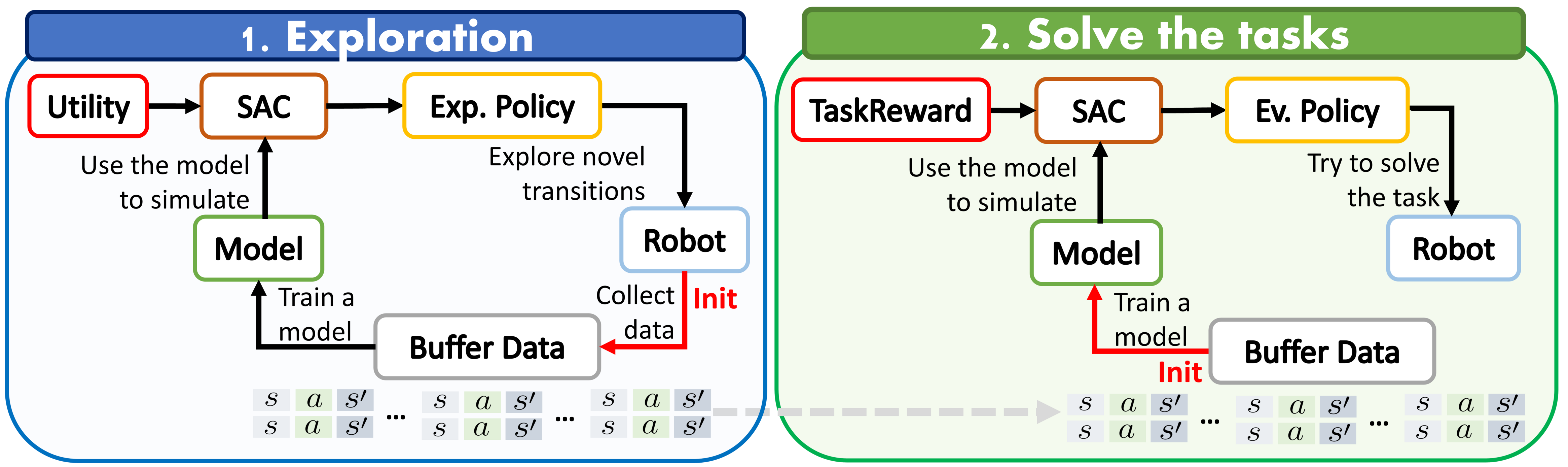}
    \caption{Summary of both exploration and evaluation pipelines with their key stages. The starting points are marked with {\color{red} Init}. Particularly, the exploration begins collecting data from a random policy and the evaluation starts from the buffer of trajectories collected along the exploration.}
    \label{fig:Pipeline}
\end{figure}

\section{PIPELINE}
The pipeline followed to explore and evaluate is inspired by Shyam et al. \cite{MAX}. 
This section describes our pipeline main stages, which are summarized in Figure~\ref{fig:Pipeline}.

\subsection{Exploration Pipeline} The exploration algorithm may be summarized as follows: \\
1) The agent starts acting randomly along $n^{ex}_{warm}$ steps. As our agent interacts with the environment, it collects trajectories of the form $\{s_{i}, a_{i}, s_{i}'\}$ in the buffer $\mathcal{D}$, where $i$ denotes the number of exploration steps. \\
2) We train a model $f^{ex}$ from $\mathcal{D}$ that learns to simulate the dynamics of the robot. We use this model and the information gain measure as utility to build an exploration policy $\pi^{ex}$.\\
3) The agent acts along $n_{pol}$ steps following policy $\pi^{ex}$. We continue saving these steps $\{s_{i}, a_{i}, s_{i}'\}$ in $\mathcal{D}$. 

We repeat stages 2-3 until the agent reaches $n^{ex}_{steps}$ exploration steps. Meanwhile, each $n_{eval}$ exploration steps, we evaluate the exploration accomplished $\mathcal{D}$ up to that moment. 

\subsection{Evaluation Pipeline} We evaluate the exploration conducted by attempting to solve a task using the knowledge gained so far, though this step is unnecessary in the real system. 
Three-step phase: \\
1) For each task of the environment, we repeat $n_{k}$ times the evaluation in order to get a more accurate estimate of the agent's performance. In each of the $n_{k}$ evaluations, the state space is randomly initialized, encouraging generalization. \\
2) We train a model $f^{ev}$ with all the collected trajectories $\mathcal{D}$  to simulate the dynamics of the robot. We use this model to build an evaluation policy $\pi^{ev}$ following the task reward.  \\
3) The agent acts along $n^{ev}_{steps}$ steps following policy $\pi^{ev}$. We compute the evaluation metric from the task reward.  \\

For learning both pure exploration and task-specific policies, we employed Soft-Actor Critic (SAC), which is the gold standard for RL in robotics \cite{haarnoja2018soft}. Specifically, we run $n^{pol}_{eps}$ episodes along $n^{pol}_{steps}$ steps, during which the trained model $f$ forecasts the consequences of the taken actions. 

This offers a significant efficiency advantage over model-free methods, as the model simulates most steps. For instance, model-free SAC would require $n_{t}\cdot n_{k} \cdot n^{SAC}_{steps}$ interactions ($n_{t}$ tasks), whereas our pipeline only executes $n^{ex}_{steps} \ll n^{SAC}_{steps}$ (Section~\ref{Sec:Results}).

\section{Evaluation details}\label{Sec:Results}
\subsection{Environments}
The evaluation of our approach to model-based active exploration is run on two benchmarks: OpenAI Gym \cite{brockman2016openai} and RLBench \cite{james2020rlbench}. The tasks of OpenAI Gym are built using the MuJoCo physics engine \cite{todorov2012mujoco}. These are some commonly used environments in the RL literature, and we use them for comparison with other studies. RLBench is built around \texttt{CoppeliaSim}\cite{rohmer2013v}. This is a complete robotic simulator with a wide range of robotic manipulation end-tasks, increasing the task's difficulty and consequently amplifying the challenge for the exploration algorithm. It properly models collisions, interactions and grasping phisics, making it much harder to grasp and manipulate objects compare to standard robotic benchmarks in RL (e.g.: robotics gym). To the best of our knowledge, we are the first to test exploration metrics and BDL models using MBRL in this simulator. Specifically, the environments and tasks we use are the following ones: \\
\textbullet{}  \texttt{HalfCheetah}-. It is a MuJoCo based planar kinematic string ($|\mathcal{S}|=17, |\mathcal{A}|=6$) 
in which we considered two different tasks: (1) \textit{Running}: move forward as fast as possible and (2) \textit{Flipping}: perform flips. \\
\textbullet{}  \texttt{Panda Robot}-. We utilize the Panda robot from Franka Emika within CoppeliaSim ($|\mathcal{S}|=43, |\mathcal{A}|=8$). The action space is directly the joint commands, therefore the model should also learn the robot kinematics and dynamics, compared to previous benchmarks with worked in Cartesian coordinates of the end-effector \cite{sancaktar2022curious}.
We employed two different RlBench tasks: (1) \textit{PushButton}: approximating its grip towards a button and (2) \textit{MoveBlockToTarget}: push a red block positioned stochastically on a table towards the goal. On success, both rewards are $100$; on failure, a non-zero value reflects the proximity to success.

\subsection{Experimental details}
The base neural network consists of a multi-layer perceptron (MLP) with $5$ layers and $512$ neurons per hidden layer. The input and output sizes of the MLP are $|\mathcal{S}|+|\mathcal{A}|$ and $2|\mathcal{S}|$, respectively. We build the Bayesian approaches taking $N=32$ models/forward passes/samples. Dropout uses probability $p=0.25$ in the middle hidden layer, and Laplace is applied only to the $n_{subnet}=1k$ highest-magnitude weights.

Concerning the pipeline, we run $n^{ex}_{steps}=20k$ exploration steps (the first $n^{ex}_{warm}=256$ steps are warm-up). Besides, we recompute the exploration policy and evaluate each $n_{pol}=25$ and $n_{eval}=2000$ exploration steps, respectively.
First, to compute the exploration policy, we run $n^{pol}_{eps}=50$ episodes along $n^{pol}_{steps}=50$ steps where the consequence of $n_{act}=128$ actions are computed simultaneously. Second, to evaluate the specific task we run $n_k=3$ episodes for $n^{ev}_{steps}=100$ steps in which we start computing the task policy ($n^{pol}_{eps}=250, \; n^{pol}_{steps}=100, \; n_{act}=128$). 

\subsection{Baselines}
We categorize relevant baselines into three groups: \\
1) Model-free RL: Utilizing SAC \cite{haarnoja2018soft}, we construct a policy for each task with the same steps on the robotic platform as our exploration process. Besides, we set an upper bound on performance by running SAC for many number of steps. \\
2) Model-based Exploration: maintaining the pipeline, we run a reactive exploration algorithm as PERX \cite{pathak2017curiosity}. We also test a simple random exploration. \\
3) Model-based active Exploration (MAX) \cite{MAX}: this work calculates the Jensen-Rényi divergence with a 32-deep ensemble based on equation \eqref{eq:js_sample}. \\ 
4) Curiosity driven exploration \cite{sancaktar2022curious} also uses ensembles and a variant of our entropy metric based on ensemble disagreement; although focused on world models, its strategy adapts to our context, and we call it \emph{EnsEnt}.

\subsection{Our strategies}
As commented before, we focus in Laplace and MC-dropout methods, with different configurations and metrics:\\
1) \emph{LapMCEnt:} This method uses the Laplace approximation in a subnetwork as described in Section \ref{sec:approxbayes}, where the predictive posterior is approximated using MC sample from the Gaussian posterior of the model. We also use the entropy metric from Section \ref{sec:entropy}.\\
2) \emph{LapMCRényi:} This method uses the Laplace approximation as before and the Jensen-Rényi divergence metric.\\
3) \emph{MCdropEnt:} This method used Monte Carlo dropout as described in Section \ref{sec:approxbayes} and the entropy metric from Section \ref{sec:entropy}.\\
4) \emph{MCdropRényi:} Finally, this method uses the MC dropout and the Jensen-Rényi divergence metric.

We also tried Laplace with a linearized Gaussian predictive posterior, but its cost was comparable to ensembles, and Laplace on the last layers only, but performance decreased; we omit both for clarity.

\begin{figure}[!t]
    \centering
    \includegraphics[width=0.95\linewidth]{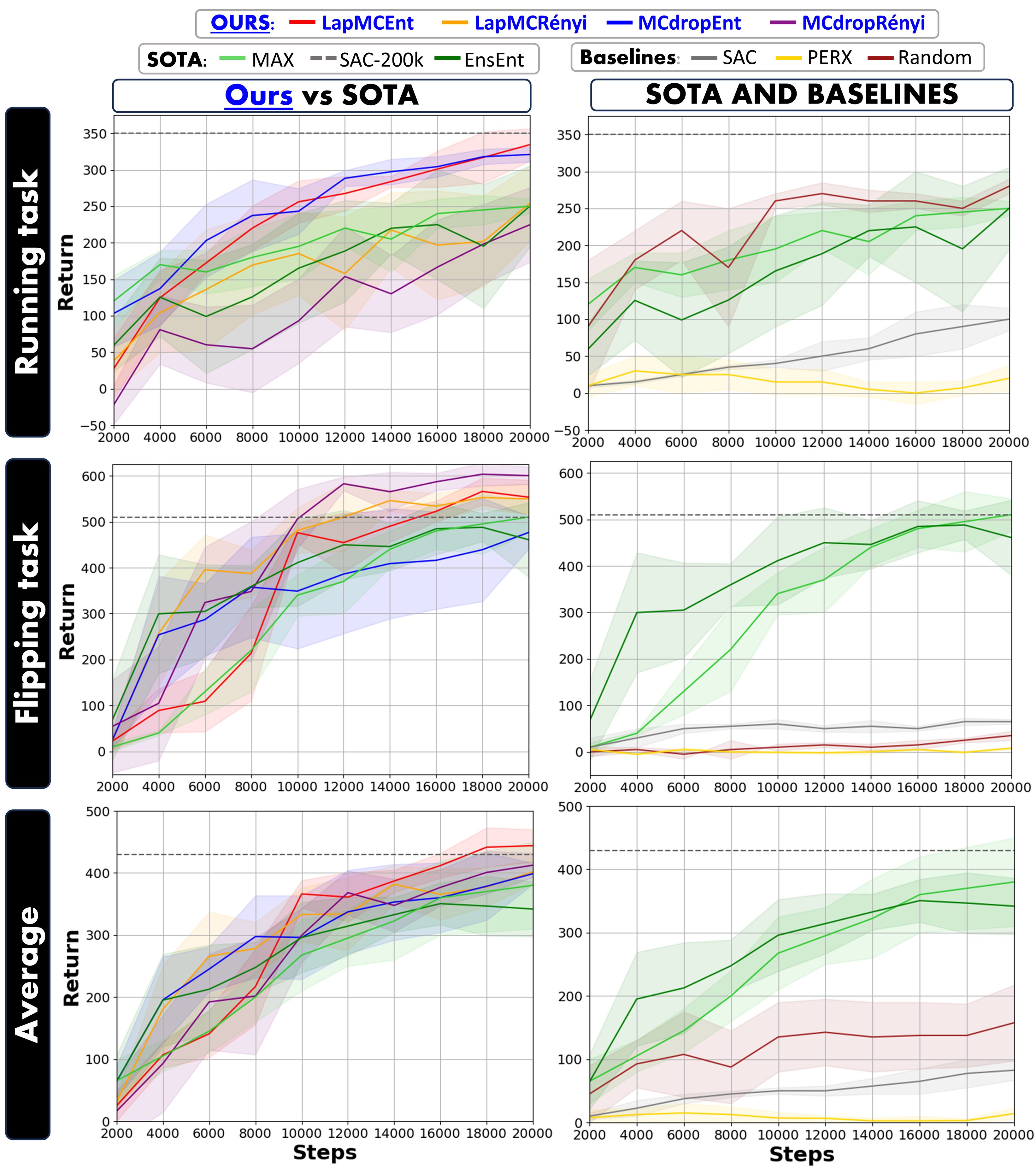}
    \caption{Results in \texttt{HalfCheetah}. First column model-based active exploration algorithms  (\textbf{\underline{\textcolor{blue}{Ours}} vs}  MAX). Second column, other approaches as presented in \cite{MAX}. MAX and SAC trained for 200k steps are represented in both sides for reference. 
    }
    \label{fig:HalfCheetah}
\end{figure}

\begin{figure*}[!t]
    \centering
    \includegraphics[width=.98\linewidth]{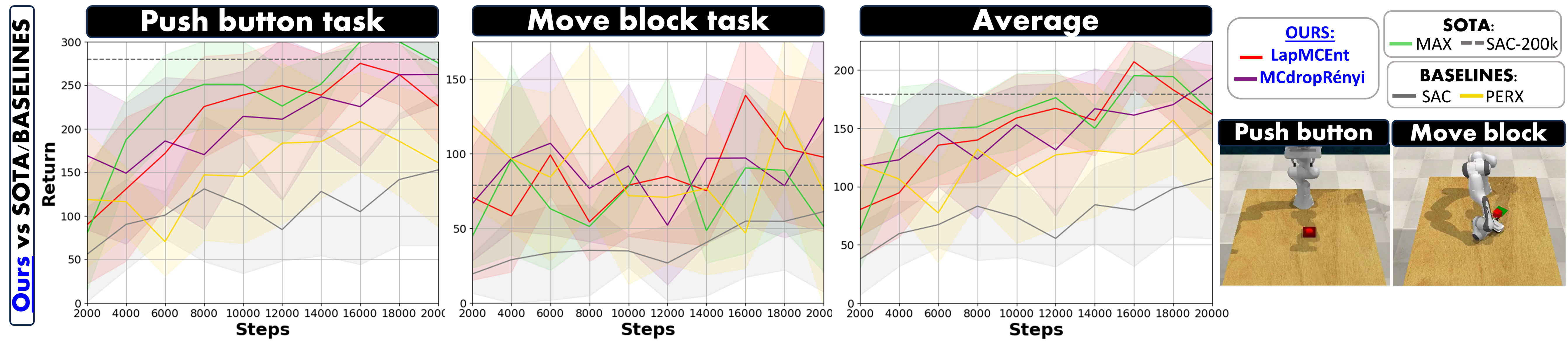}
    \caption{Results in \texttt{Panda Robot} for \textbf{\underline{\textcolor{blue}{our best approaches}}} vs SOTA MAX and baselines. The Laplace approximation with our entropy metric is more robust in the more difficult tasks (move block) due to its better support, while sampling methods overfit to straightforward tasks (push button).}
    \label{fig:Coppelia}
\end{figure*}

\section{RESULTS}
In this section we compare the Bayesian models and utility measures in terms of reward, computational time and storage. 

\subsection{Rewards}\label{Sec:rewards}
We quantify performance by examining the rewards achieved during the evaluation phase, understanding that a higher task reward is an indicative of effective exploration. To ensure a fair comparison, we employ the same Bayesian model in the evaluation stage, uniquely varying the buffer (collected along the exploration) used for model training. 

\textbf{HalfCheetah}-.  Figure~\ref{fig:HalfCheetah} shows how Bayesian model-based active exploration approaches clearly outperform the remaining methods. Specifically,  approaches where the Bayesian model is built from Laplace approximation and MC-dropout (ours) achieve higher performance than those built from deep ensembles. Besides, to analyze the impact of the utility function on overall performance, we compare the task rewards achieved using the same BDL techniques while varying the utility function (Rényi vs. our \emph{entropy metric}). These results indicate that \emph{our entropy measure} is more suitable for Laplace-based model, whereas deep ensembles fit better with Rényi entropy. 
Lastly, note that our \textit{LapMCEnt} achieves higher task rewards than baseline SAC with an order of magnitude lower training exploration steps (20k vs 2$\times$200k).

\textbf{Panda Robot}-. Leveraging the best-performing methods in \texttt{HalfCheetah}, we turn to our problem of interest: robot manipulation with a \texttt{Panda Robot}. In this high-dimensional environment, we observe again how Bayesian model-based active exploration approaches (including our \emph{LapMCEnt} and \emph{MCdropRényi})  outperform the baselines. Our methods achieve again higher task rewards than baseline SAC with an order of magnitude lower training exploration steps, while the performance is similar between them. In this case, we have not include \emph{EnsEnt} in the plots, because the performance was very similar to \emph{MAX}. 
Task complexity also plays an important role. While these methods solve the \textit{PushButton} task in nearly all the evaluation episodes, they have some difficulties with \textit{MoveBlock} task. The reason is that the randomization of the block position plays an important role forcing to learn robust policies and performance measurement is noisy. However, model based approach are able to exploit model uncertainty and learn robust policies even with few data points, that is why the plots tend to be flat for Bayesian methods and gathering more data barely improve the performance. 
Note how the performance of Bayesian methods is comparable to SAC-200k with less than 2000 data points. 
Particularly, the peak in the average of both tasks is reached by \textit{LapMCEnt}.

\subsection{Calibration}
On \texttt{HalfCheetah}, we also assess calibration, i.e., how well the prediction error aligns with the model's uncertainty estimation, using the Area Under Sparsification Error (AUSE) metric \cite{gustafsson2020evaluating}. We train each Bayesian model on the first 18k steps of the exploration buffer and test on the last 2k, comparing prediction errors against their utility values. As Table \ref{tab:calibration_time} shows, the Laplace-based model with our \emph{entropy metric} achieves the highest calibration, while MC-dropout and deep ensembles with Rényi entropy (MAX) trail further.

\begin{table}[!tb]
    \centering
    \caption{Calibration and computational time metrics. Best in bold.}
    \label{tab:calibration_time}
    \begin{tabular}{| c | c | c  c |}
    \hline
        Model & AUSE $\downarrow$ & Training (s) $\downarrow$ & Inference (s) $\downarrow$ \\
        \hline
        \textit{MAX} \cite{MAX} & 0.075 & 122.75 & 6.25e-3 \\
        \hline
        \arrayrulecolor{black}
        \textit{LapMCEnt} (Ours) &  \textbf{0.048} & 39.98 & \textbf{6.00e-3} \\
        \textit{MCdropRényi} (Ours) & 0.085 & \textbf{34.35} & 6.91e-3 \\
        \hline
    \end{tabular}
\end{table}

\subsection{Computational time and load}
Table \ref{tab:calibration_time} also compares the models in computational time and memory. Implementing MAX increases training time, as it requires fitting $n_{ens}$ neural networks; the Laplace Approximation adds some overhead over MC-dropout, since it must fit the posterior covariance \eqref{eq:LapApp}, while inference times show no substantial differences.

The storage cost depends on the number of model weights ($n_{weights}$), the number of networks stacked for deep ensembles ($n_{ens}$), and the covariance matrix size of the Laplace Approximation ($n_{subnet}^{2}$):
 \begin{equation}
     Cost = \mathcal{O}\left(n_{ens} \cdot n_{weights} + n_{subnet}^{2}\right).
 \end{equation}
In our scenario, wherein our MLP encompasses $n_{weights} \approx 1M$ of weights, opting for deep ensembles (MAX) leads to a memory overhead 32 times greater than that of MC-dropout and 16 times greater than that of Laplace Approximation. 

In summary, our methods match or exceed MAX while greatly reducing training time and memory, and \emph{LapMCEnt} further improves calibration.

\section{CONCLUSIONS}
This work demonstrates the benefits of model based RL for robotics systems, where sample efficiency is critical, by replacing expensive real world interactions by cheap model evaluations during policy training, and by performing active learning on the model to further reduce the need for real world interactions by selecting the most informative ones. In addition, once trained, the task-agnostic model can be reused to solve multiple tasks. We show that the extended use of deep ensembles hides the actual issues of lack of calibration and limits applicability due to its high computational and memory footprint, whereas our proposed Bayesian models —based on Laplace approximation and MC-dropout— offer enhancements in performance, calibration, and efficiency while paving the way for novel utility formulations.

\bibliographystyle{unsrt} 
\bibliography{bibliography.bib}

\end{document}